# Utilisation des grammaires probabilistes dans les tâches de segmentation et d'annotation prosodique

*Irina Nesterenko[1], Stéphane Rauzy[2]*

[1]Université Blaise Pascal, Clermont-Ferrand II
[2]Laboratoire Parole et Langage, CNRS & Université de Provence
irina.nesterenko@univ-provence.fr, stephane.rauzy@lpl-aix.fr

**ABSTRACT**

Methodologically oriented, the present work sketches an approach for prosodic information retrieval and speech segmentation, based on both symbolic and probabilistic information. We have recourse to probabilistic grammars, within which we implement a minimal hierarchical structure. Both the stages of probabilistic grammar building and its testing in prediction are explored and quantitatively and qualitatively evaluated.

## 1. INTRODUCTION

Dans la présente contribution, nous explorons la question du rôle des informations mélodiques et probabilistes dans le processus de segmentation du continuum sonore en unités correspondant aux mots. La problématique de la segmentation du signal continu attire l'attention à la fois des psycholinguistes et des spécialistes en technologies de la parole, notamment en reconnaissance vocale ou en annotation automatique des corpus oraux. En mettant en parallèle ces deux domaines de recherche, nous soutenons l'idée, d'après laquelle la performance des applications technologiques s'améliorera si les modèles implémentés s'appuient sur les mécanismes utilisés par des humains dans des tâches similaires.

Ainsi, il a été démontré dans les études psycholinguistiques que le processus de la segmentation du signal continu de parole en mots est largement guidé par les informations phonétiques fines à granularité variable. Il s'ensuit que les indices acoustiques de la segmentation du continuum sonore sont redondants. Les auditeurs s'appuient sur les variations allophoniques, sur les patrons phonotactiques [1], sur les contraintes métriques [2-3] ou encore sur les patrons temporels et mélodiques [4-5]. Cependant, il est nécessaire d'insister sur leur rôle non-déterministe : ces indices acoustiques suggèrent l'emplacement d'une frontière mais ne le déterminent pas. Si le rôle des indices tonals a été démontré pour le français, ceci n'est pas le cas pour la langue russe.

Notre étude s'appuie sur un corpus de parole spontanée en langue russe. Nous considérons que les corpus annotés sont une ressource linguistique inestimable dans la mesure où ils nous renseignent sur le fonctionnement du système langagier en perception et en production. En technologies de la parole, faire appel aux techniques de corpus permet de prendre en compte la dimension de la variabilité propre au discours humain. Il est par conséquent nécessaire de disposer de grands volumes de données annotées et des heuristiques semi-automatiques facilitant la démarche d'annotation. En même temps, améliorer la performance des outils d'annotation automatique sert également d'outil de vérification des modèles linguistiques sous-jacents.

La recherche que nous avons menée a un double intérêt : d'une part, nous nous penchons sur la question de la contribution des informations tonales au marquage d'unités prosodiques mineures en langue russe ; d'autre part, dans la quête d'une heuristique semi-automatique pour prédire les frontières de mots, nous évaluerons le rôle d'une représentation hiérarchique de la constituance prosodique.

Cet article est organisé comme suit : dans la section 2, nous présentons le plan expérimental de l'étude, en nous arrêtant, en particulier, sur les questions de l'annotation des phénomènes mélodiques et de la structure prosodique ; nous décrivons également le corpus d'étude et l'appareil mathématique mis en œuvre. Les principaux résultats en relation avec l'étape de la construction des modèles probabilistes et celle des tests en prédiction sont exposés en section 3. Enfin, la section 4 est dédiée à des conclusions et à la discussion générale.

## 2. DESIGN EXPÉRIMENTAL

### 2.1. Modèle prosodique

Nous interprétons le système prosodique comme étant un système organisateur entièrement spécifié à travers les représentations formelles des paliers métrique, tonal et celui du phrasé prosodique.

Pour le niveau tonal, nous avons adopté le système d'annotation Momel-Intsint [6]. La représentation prosodique associée à ce schéma d'annotation suppose l'existence de trois niveaux intermédiaires entre le signal de parole et la représentation fonctionnelle : on distingue le niveau phonétique, le niveau phonologique de surface et le niveau phonologique profond. Le niveau phonétique correspond à une représentation continue, reflétée par la courbe f0 après la soustraction de la composante microprosodique [7]. Cette représentation est obtenue moyennant l'application de l'algorithme Momel. La modélisation se fonde sur la détection des points cibles de la courbe mélodique, ces dernières étant interpolées par des fonctions de splines quadratiques. Par conséquent, cette modélisation est dans un premier temps acoustique. En même temps, il semble légitime d'interpréter les points cibles comme étant des sites où le locuteur modifie le mouvement de la f0 de manière volontaire afin de parvenir à ses intentions communicatives.

Au niveau phonologique de surface, les points cibles reçoivent l'encodage symbolique en termes d'alphabet Intsint. Le système INTSINT comprend un alphabet de huit symboles {T(op), M(id), B(ottom), H(igher), S(ame), L(ower), U(pstepped), D(ownstepped)} qui se répartissent en deux grands groupes : les tons absolus (T, M, B) réservés au codage des valeurs définissant l'étendue tonale du locuteur (cet encodage s'appuie sur deux paramètres, clef et étendue) et les tons relatifs (H, S, L, D, U) définis en référence aux points cibles adjacents. Les symboles relatifs se divisent à leur tour en deux groupes, la distinction étant faite entre les tons non-itératifs (H, S, L) et les tons itératifs (U, D). Les grammaires probabilistes de notre étude modélisent l'enchaînement des symboles Intsint dans l'encodage des courbes mélodiques associés aux mots prosodiques.

### 2.2. Structure hiérarchique implémentée

Un des objectifs de la présente étude a été de modéliser l'impact de la structure hiérarchique minimale, car les représentations formelles de la structure prosodique, développées dans le cadre de la phonologie prosodique et de la phonologie intonative, sont hiérarchisées. Chaque niveau de la constituance prosodique représente une interface entre le module phonologique et un autre module de la grammaire. En règle générale, malgré les variations terminologiques et conceptuelles entre les différentes auteurs, deux niveaux supérieurs au mot prosodique sont distingués : le niveau des syntagmes phonologiques et celui des syntagmes prosodiques. La définition des ses constituants est guidée soit par des contraintes morpho-syntaxiques, soit par les seules informations prosodiques.

Dans la présente étude, les mots prosodiques constituent le niveau inférieur dans la hiérarchie des constituants implémentée. La structure hiérarchique implémentée prend en considération un seul niveau immédiatement supérieur au niveau des mots prosodiques. En même temps, ce niveau supérieur n'appartient pas à la constituance prosodique proprement dite : il s'agit du niveau des tours de parole, donc du niveau de l'organisation discursive. Ce choix est motivé d'abord par le caractère méthodologique de notre étude ; l'impact de la structure hiérarchique plus complexe nécessite d'être étudié dans des recherches postérieures.

Une question annexe est celle du lien entre un mot prosodique et un mot lexical. Le mot prosodique est traditionnellement défini comme incluant sous le même nœud le mot lexical et tous ses clitiques. Or, les études en production démontrent que ce sont les mots prosodiques qui sont les unités d'encodage phonologique à l'étape de la conceptualisation de l'énoncé [8].

### 2.3. Corpus et mise en forme des données

L'étude empirique s'appuie sur un corpus de parole spontanée en langue russe (recueilli dans le cadre du projet Intas-915). Nous avons analysé la production d'une locutrice native de la langue russe dans la conversation informelle (temps de parole 17 minutes pauses comprises).

Le corpus a été manuellement annoté en tours de parole et en mots prosodiques et ultérieurement, chaque mot prosodique a été annoté selon qu'il portait ou non la proéminence mélodique. Notre corpus comprend 825 mots prosodiques.

À l'étape suivante, les patrons tonals associés à chaque mot ont été extraits. Les annotations réalisées sous Praat ont été ensuite converties au format XML (Figure 1). Cette représentation constitue les données d'entrée pour le module de construction des grammaires probabilistes qui permettent de découvrir les régularités dans les séquences d'étiquettes tonales Intsint. Dans la section suivante, nous présentons l'appareil mathématique sur lequel repose notre étude.

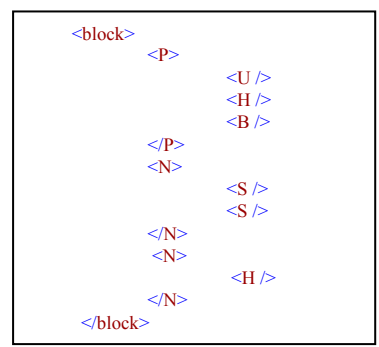

**Figure 1 :** Format de données

### 2.4. Appareil mathématique

La présente étude repose sur la présupposition qu'il existe des régularités dans l'enchaînement des étiquettes Intsint, ces régularités dépendant elles-mêmes de la réalisation de telle ou telle primitives fonctionnelles. Pour découvrir ces régularités et les dépendances entre les étiquettes tonales, nous avons eu recours aux grammaires probabilistes. Dans cette perspective, deux concepts mathématiques sont d'intérêt : celui des probabilités conditionnelles et celui de l'entropie.

Dans le cas général, nous disposons de $N$ catégories $c_i$ qui décrivent un phénomène langagier. La distribution de ces catégories et les rapports de dépendance entre elles sont étudiées à partir d'un corpus d'entraînement. La tâche est alors d'estimer la probabilité de chaque séquence de catégories dans le corpus : dans notre cas, cette séquence de catégories correspond à la séquence des étiquettes Intsint associée à un mot prosodique, par exemple, la séquence (U, S, T, D). La probabilité de cette séquence est calculée selon la formule

$$P(c_1, c_2, c_3, c_4) = \pi_1 * \pi_2 * \pi_3 * \pi_4,$$

où $\pi_1 = P(c_1)$, $\pi_2 = P(c_2| c_1)$, $\pi_3 = P(c_3| c_1, c_2)$ et $\pi_4 = P(c_4| c_1, c_2, c_3)$. La valeur $\pi_3 = P(c_3| c_1, c_2)$ est la probabilité conditionnelle de la catégorie $c_3$ étant donnée qu'elle est précédée par la séquence d'étiquettes tonales ($c_1, c_2$).

Dans le calcul des probabilités, nous utilisons le modèle des patrons [9], un modèle de Markov caché caractérisé par une bonne extraction de l'information contenue dans le corpus d'entraînement. Contrairement aux modèles de type n-grammes, le contexte gauche des patrons du modèle (les états de l'automate à états finis) n'est pas limité à un nombre fixe de symboles mais est de longueur variable. Si une séquence d'étiquettes est fréquente dans le corpus, le modèle calcule et mémorise ces probabilités conditionnelles pour toutes les catégories étant donné le patron attesté.

Trois modèles ont été calculés :
- **Modèle Plat** : la structure hiérarchique n'est pas prise en compte (les symboles codant les catégories se limitent aux tons et aux symboles codant l'ouverture et la fermeture des tours de parole;
- **Modèle Hiérarchique** : deux niveaux de constituance sont implémentés (il s'agit du niveau précisant les tours de parole et du sous-niveau codant les frontières des mots prosodiques) ;
- **Modèle Hiérarchie&Proéminence** : ce modèle contient à la fois la spécification de deux niveaux de constituance et la distinction entre les mots prosodiques porteurs de la proéminence mélodique *versus* les mots prosodiques non-proéminents.

Pour évaluer la performance des modèles, nous avons eu recours à la mesure de l'entropie qui quantifie l'organisation de l'information dans le système. Pour une distribution, l'entropie quantifie la différence entre cette distribution et la distribution équiprobable des catégories. L'entropie varie donc entre 0 et $ln\ N$ ($N$ étant le nombre de catégories dans le système d'annotation) : une entropie nulle caractérise un système déterministe et l'entropie maximale de $ln\ N$ – un système équiprobable. Le concept de l'entropie normalisée permet de situer les valeurs de l'entropie dans l'intervalle [0; 1]. Pour évaluer la performance des modèles de patrons, nous avons calculé l'entropie de la distribution des catégories avec et sans le modèle langagier implémenté.

Une fois les grammaires probabilistes calculées, nous avons testé leurs performances en prédiction. Étant donné la séquence d'étiquettes tonales et la grammaire probabiliste sous-jacente, l'algorithme de Viterbi est utilisé pour rechercher la distribution optimale des frontières des mots prosodiques. La performance des modèles a été quantifiée à l'aide des mesures de rappel, précision et f-mesure, traditionnellement utilisées en recherche documentaire.

## 3. RÉSULTATS

### 3.1. Calcul des modèles probabilistes

Dans le Table 1, nous présentons les valeurs de l'entropie et de l'entropie normalisée pour trois modèles : Modèle Plat, Modèle Hiérarchique et Modèle Hiérarchie&Proéminence. SansModèle correspond à la mesure de l'organisation de l'information si aucun modèle de patrons n'est spécifié. Les valeurs observées en absence des modèles langagiers sont proches de l'unité pour l'entropie normalisée, indiquant ainsi que la distribution des catégories utilisées (les tons et les symboles codant les frontières de mots) est proche d'une distribution équiprobable. Pour le cas Hiérarchique, on observe que l'entropie est légèrement plus basse, ce qui résulte du fait que les symboles codant l'ouverture et la fermeture d'un mot prosodique sont plus nombreux que les autres catégories. Les valeurs de l'entropie et de l'entropie normalisée diffèrent significativement lorsque les modèles sont pris en compte. Qui plus est, l'effet de conditionnement est plus marqué dans le cas où la structure hiérarchique minimale est considérée.

**Table 1 :** Entropie et entropie normalisée de trois modèles probabilistes

| Modèle | Entropie | | Entropie normalisée | |
|---|---|---|---|---|
| | SansM | AvecM | SansM | AvecM |
| Plat | 2.259 | 1.796 | 0.942 | 0.749 |
| Hiérarchique | 2.064 | 1.494 | 0.897 | 0.649 |
| H&Proéminence | 2.696 | 1.638 | 0.915 | 0.556 |

Comme nous l'avons spécifié précédemment pour rendre compte des événements mélodiques fonctionnels et contrastifs, l'annotation Intsint nécessite d'être appariée avec l'annotation des fonctions prosodiques (IF). Nos résultats montrent qu'en effet, il existe des régularités dans l'enchaînement d'étiquettes Intsint et que ces régularités sont plus finement modélisées lorsque l'information apportée par les catégories fonctionnelles est prise en compte.

## 4. TEST EN PRÉDICTION

Dans un second temps, nous avons évalué la performance du Modèle Hiérarchique et du Modèle Hiérarchie & Proéminence en prédiction : la tâche était de prédire la distribution des frontières des mots prosodiques à partir des représentations symboliques de la courbe mélodique. L'heuristique de prédiction s'appuie sur l'algorithme de Viterbi : le modèle évalue les différentes solutions avec les frontières insérées après chaque symbole Intsint. L'algorithme cherche la distribution optimale des frontières de mots prosodiques à l'intérieur du tour de parole, c'est-à-dire à l'intérieur du niveau supérieur dans la hiérarchie des constituants. Nous présentons dans les Tables 2-4 les matrices de confusion ainsi que les mesures d'évaluation des performances de modèles.

**Table 2 :** Matrice de confusion du modèle Hiérarchique

|  | Prédiction | |
|---|---|---|
|  | pas de frontière | frontière |
| pas de frontière | 2003 | 190 |
| frontière | 329 | 497 |

**Table 3 :** Matrice de confusion du modèle Hiérarchie & Proéminence

|  | Prédiction | |
|---|---|---|
|  | pas de frontière | frontière |
| pas de frontière | 2003 | 194 |
| frontière | 398 | 428 |

**Table 4 :** Métriques d'évaluation

|  | Modèle Hiérarchique | Modèle Hiérarchie&Proéminence |
|---|---|---|
| Précision | 0.72 | 0.688 |
| Rappel | 0.6 | 0.518 |
| F-mesure | 0.655 | 0.591 |

La mesure de rappel quantifie la proportion des prédictions correctes. Le modèle Hiérarchique bénéficie d'un rappel de 0,6 et le modèle Hiérarchie&Proéminence d'un rappel de 0,52. La mesure de précision quantifie le nombre de frontières correctement prédites sur le nombre de frontières insérées par l'algorithme. Dans notre étude, les valeurs de précision sont supérieures à celles du rappel : nos modèles ont ainsi tendance à insérer plus de frontières de mots que nécessaire. La performance des modèles est dépend finalement du couple <rappel, précision>, ce que reflète la F-mesure. Le modèle Hiérarchique affiche une valeur de F-mesure de 65,5%, laquelle est supérieure à celle du modèle Hiérarchie&Proéminence (59,1%).

Notre objectif principal était de tester l'impact des informations probabilistes dans la tâche de segmentation de la parole. Dans cette perspective, les résultats semblent assez concluants. La performance des modèles pourra de plus être améliorée, notamment, en implémentant une annotation fonctionnelle et une hiérarchie des constituants prosodiques plus nuancées et plus complexes.

## 5. DISCUSSION ET CONCLUSIONS

Dans la présente étude, nous avons étudié le rôle des informations tonales et probabilistes pour la segmentation en mots du signal continu de parole. L'objectif de notre étude était a) de déterminer s'il existe des régularités dans l'enchaînement des étiquettes Intsint, b) de modéliser comment les informations probabilistes dans l'espace tonal peuvent être explorées dans une tâche de segmentation du continuum sonore à la fois par les humains et par des algorithmes d'annotation semi-automatique de corpus, c) de tester si l'implémentation d'une structure hiérarchique minimale améliore la performance de l'algorithme.

Dans un premier temps, nous avons démontré qu'il existe des régularités dans les séquences des symboles Intsint associés aux mots prosodiques. Ce résultat est à mettre en perspective avec les études en psycholinguistique soulignant le rôle des informations tonales et probabilistes aux étapes initiales du traitement perceptif du signal de parole. Le rôle des informations tonales a été démontré pour le français ; nous avons démontré que ces informations peuvent être explorées par les locuteurs du russe.

Dans un deuxième temps, l'algorithme proposé a été testé en prédiction. Le modèle Hiérarchique arrive à 60% de prédictions correctes de frontières de mots prosodiques (F-mesure de 65,5%). La performance du modèle est satisfaisante. Remarquons cependant que le modèle a été entraîné sur un corpus limité : nous pouvons donc espérer que l'augmentation du volume de données annotées permettra de raffiner les grammaires probabilistes et d'améliorer ainsi la performance des modèles en prédiction.

Nous avons également constaté qu'implémenter la structure hiérarchique contribue à l'organisation de l'information à l'intérieur du système. Dans les recherches futures, nous envisageons d'étudier par la méthode des grammaires probabilistes l'impact des annotations plus détaillées du phrasé prosodique et des fonctions prosodiques.

## BIBLIOGRAPHIE

[1] Weber, A. The role of phonotactics in the segmentation of native and non-native continuous speech. In *Proc. Workshop on Spoken Access Processes*, 143-146, 2000.

[2] Cutler, A., Mehler, J., Norris, D. and Segui, J. The syllable's differing role in the segmentation of French and English. *Journal of Memory and Language*, Vol. 25, 1986, 385-400.

[3] Cutler, A., Otake, T. Mora or phoneme? Further evidence for language-specific listening. *J. Exp. Psych.: Human perception and performance,* Vol. 14, 1994, 113-121.

[4] Turk, A.E., Shattuck-Hufnagel, S. Word-Boundary-Related Duration Patterns in English. *J. of Phonetics,*

Vol. 28, 2000, 397-440.

[5] Di Cristo, A. Vers une modélisation de l'accentuation du français : seconde partie. *French Language Studies*, Vol. 10, 2000, 27-44.

[6] Hirst, D., Di Cristo, A., Espesser, R. Levels of description and levels of representation in the analysis of intonation. In M. Horne (ed.), *Prosody: Theory and Experiment*, Dordrecht : Kluwer Academic Press, 51-87, 2000.

[7] Di Cristo, A., Hirst, D. Modelling French micromelody: analysis and synthesis. *Phonetica*, 43 (1):11-30, 1986

[8] Levelt, W. J. M. Accessing words in speech production: stages, processes and representations. *Cognition*, 42, 1992, 1-22.

[9] Blache P., Rauzy, S. Mécanismes de contrôle pour l'analyse en Grammaires de Propriétés. In *Proc. TALN Conference*, 2006, pages 415-424.